\documentclass[11pt]{article}
\ifdefined\arxiv
  \pdfoutput=1
\fi

\usepackage[final]{acl}

\usepackage{times}
\usepackage{latexsym}
\usepackage{url}
\usepackage{siunitx}

\usepackage[T1]{fontenc}
\usepackage[utf8]{inputenc}

\usepackage{microtype}

\usepackage{inconsolata}
\usepackage{booktabs}
\usepackage{graphicx}

\usepackage{tcolorbox}
\tcbuselibrary{breakable}

\newtcolorbox{promptbox}{
    colframe=gray,
    boxrule=0.5pt,
    colback=white,
    fontupper=\ttfamily\small,
    breakable
}

\title{ELR-1000: A Community-Generated Dataset for Endangered Indic Indigenous Languages}
\author{
    \parbox{0.95\linewidth}{
        \centering{
            $^\star$ Neha Joshi\textsuperscript{\normalfont $\diamondsuit$} \quad
            $^\star$ Pamir Gogoi\textsuperscript{\normalfont $\diamondsuit$} \quad 
            Aasim Mirza\textsuperscript{\normalfont $\diamondsuit$} \quad
            Aayush Jansari\textsuperscript{\normalfont $\diamondsuit$} \\
            Aditya Yadavalli\textsuperscript{\normalfont $\spadesuit$} \quad
            Ayushi Pandey\textsuperscript{\normalfont $\diamondsuit$} \quad
            Arunima Shukla\textsuperscript{\normalfont $\diamondsuit$} \quad
            Deepthi Sudharsan\textsuperscript{\normalfont $\clubsuit$} \\
            Kalika Bali\textsuperscript{\normalfont $\heartsuit$} 
            \quad
            Vivek Seshadri\textsuperscript{\normalfont $\diamondsuit$} \\
            {\rm \textsuperscript{\normalfont $\diamondsuit$} Karya \quad \textsuperscript{\normalfont $\spadesuit$} UC San Diego \quad 
            \textsuperscript{\normalfont $\clubsuit$} Independent Researcher
            \textsuperscript{\normalfont $\heartsuit$} Microsoft Corporation}\\
            {\tt \{neha, pamir.gogoi, vivek\}@karya.in}
        }
    }
}

\usepackage{lipsum}

\newcommand\blfootnote[1]{%
  \begingroup
  \renewcommand\thefootnote{}\footnote{#1}%
  \addtocounter{footnote}{-1}%
  \endgroup
}

\begin{document}
\maketitle
\begin{abstract}
We present a culturally-grounded multimodal dataset of 1,060 traditional recipes crowdsourced from rural communities across remote regions of Eastern India, spanning 10 endangered languages. These recipes, rich in linguistic and cultural nuance, were collected using a mobile interface designed for contributors with low digital literacy. Endangered Language Recipes (ELR)-1000---captures not only culinary practices but also the socio-cultural context embedded in indigenous food traditions. We evaluate the performance of several state-of-the-art large language models (LLMs) on translating these recipes into English and find the following: despite the models' capabilities, they struggle with low-resource, culturally-specific language. However, we observe that providing targeted context---including background information about the languages, translation examples, and guidelines for cultural preservation---leads to significant improvements in translation quality. Our results underscore the need for benchmarks that cater to underrepresented languages and domains to advance equitable and culturally-aware language technologies. As part of this work, we release the ELR-1000 dataset to the NLP community, hoping it motivates the development of language technologies for endangered languages.\footnote{\url{https://github.com/karya-inc/elr-1000}}
\blfootnote{$\spadesuit$ Work done while in Karya}
\blfootnote{$\clubsuit$ Work done while in Microsoft}
\blfootnote{$\star$ Joint first authors}
\end{abstract}

\section{Introduction}

Indian natural language processing communities have seen a renewed wave of digitization of constitutional languages \cite{10.1145/3290605.3300611, abraham-etal-2020-crowdsourcing, madaan2022anuvaad, doddapaneni-etal-2023-towards, khan-etal-2024-indicllmsuite, 10.1145/3663775}. However, most datasets focus on high-resource or officially recognized languages, overlooking India's tribal and indigenous diversity. This exclusion limits access to digital tools and information, deepening inequalities and endangering linguistic heritage. %Many of these languages carry vital oral knowledge in areas like agriculture and medicine, making their loss a threat to cultural and ecological memory.

%However, these datasets overwhelmingly focus on high-resource or constitutionally-recognized languages, failing to account for the rich linguistic diversity of India's indigenous and tribal communities. 
Due to this gap, speakers of non-standardized, underrepresented, or endangered languages remain excluded from digital services, and educational resources. This technological gap compounds existing inequalities and accelerates the obsolescence of their languages. Such marginalization contributes not only to cultural erasure but also to the loss of generational knowledge---particularly in domains like agriculture, medicine, and food systems, where knowledge is embedded in oral traditions. These languages are repositories of unique cultural epistemologies, and their extinction implies the loss of irreplaceable linguistic and ecological knowledge. Incorporating these low-resource languages into mainstream NLP systems can help bridge information access gaps, supporting community agency, digital inclusion, and long-term linguistic preservation.

\begin{table*}[t]
\centering
\small
\setlength{\tabcolsep}{3pt}
\renewcommand{\arraystretch}{1.2}
\resizebox{\textwidth}{!}{
\begin{tabular}{@{}p{3.3cm}p{1.8cm}p{2.5cm}p{2.3cm}p{2.8cm}p{3cm}p{2.5cm}@{}}
\toprule
\textbf{Dataset} & \textbf{Region} & \textbf{\# Entries} & \textbf{Modalities} & \textbf{Main Data Source} & \textbf{Contextual Info} & \textbf{Lang. Coverage} \\
\midrule
Recipe1M+~\citep{marin2021recipe1m+} & Global & 13M images & Text, Image & Web & None & English \\
IndianFoodNet~\citep{agarwal2023indianfoodnet} & India & 5,500+ images & Image & Web & None & English \newline / Indic labels \\
Khana~\citep{prabhu2025khana} & India & 131K images & Image & Web & Visual features & English \\
ISIA Food-500~\citep{min2020isia} & Global &  399,726 images & Image & Web & None & English \\
\textbf{ELR-1000 (This work)} & India & 1,060 recipes & \textbf{Text, Image, \newline Audio} & \textbf{Community} & \textbf{Seasonality, \newline storage, ecology, \newline tools, oral narration} & 10 endangered \newline Indian languages \\
\bottomrule
\end{tabular}
}
\caption{Comparison of major food computing datasets and the proposed Endangered Indic Indigenous Recipes (ELR-1000) dataset}
\label{tab:food-datasets}
\end{table*}

From a technical perspective, neural machine translation (NMT) for low-resource languages has historically underperformed due to limited parallel corpora and limited annotated data. Recent advances, however, have shown promise. Large language models (LLMs) have begun to show significantly improved performance even in zero-shot or few-shot translation settings. Models like GPT 4o \cite{openai2024gpt4ocard}, LLaMa \cite{touvron2023llama} and Gemini \cite{comanici2025gemini25pushingfrontier} demonstrate emergent capabilities in translating low-resource languages. The development of domain-specific \cite{DBLP:journals/corr/abs-2402-15061} or culturally relevant corpora \cite{yakhni-chehab-2025-llms} further enhances translation quality, making such initiatives doubly impactful.

%In this paper, we present the first systematic digitization initiative for extremely low-resourced, marginalized languages from the regions of Jharkhand, Eastern Bihar, and North-East India. Despite the linguistic richness of these regions, very little effort has been directed toward the development of computational resources for their languages. 

Our project documents 1,060 traditional recipes across 10 tribal languages, collected through a community-based approach from 368 rural women and 26 men who are native speakers of these languages. This dataset, grounded in indigenous culinary knowledge, not only contributes to linguistic resource development but also serves as a cultural archive for future generations. Most importantly, a subset of this data is also parallel data, where each recipe has been translated manually into English. Through this work, we aim to demonstrate a replicable model for ethical data collection, annotation, and deployment of AI tools for severely underrepresented language communities.

%While recent efforts have expanded NLP resources for low-resource languages, many of these datasets rely on content originally produced in high-resource languages and later translated into the target language. For instance, Flores-101 evaluation benchmark \cite{goyal-etal-2022-flores} provides translations in 101 low-resource languages, but its source content is drawn entirely from English Wikipedia, leading to a lack of cultural grounding in the target communities. In contrast, our data set is generated by speakers of low-resource and tribal languages, using content rooted in their daily lives, particularly food practices. This makes it one of the few genuinely community-authored datasets in the space, offering not just linguistic material but also cultural and epistemic representation of the communities themselves. In this paper, we present the first systematic digitization initiative for extremely low-resourced, marginalized languages from the regions of Jharkhand, Eastern Bihar, and North-East India. Despite the linguistic richness of these regions, very little effort has been directed toward the development of computational resources for their languages.

While many datasets for low-resource languages exist, they often rely on translations from high-resource languages, lacking cultural grounding. For example, the Flores-101 benchmark \cite{goyal-etal-2022-flores} covers 101 languages but is based entirely on English Wikipedia content. In contrast, our dataset is community-authored by speakers of low-resource and tribal languages, using content rooted in daily life, especially food practices. This makes it one of the few datasets that offer both linguistic and cultural representation.

The main contributions of this work are as follows:
\begin{itemize}
    \item We release ELR-1000---composed of 1,060 recipes in 10 endangered languages of Eastern Indic languages under the Karya Public License (KPL). 
    \item Translation of a representative subset of this corpus into English in the form of a parallel corpus for LLM-enabled translation.
    \item Evaluating LLM capabilities in translating traditional recipes, focusing on cultural authenticity and factual accuracy. Highlighting current strengths and limitations in handling nuanced, culturally specific content.
\end{itemize}
% These contributors come from communities where internet connectivity is sporadic, smartphone penetration is minimal, and digital literacy rates remain exceptionally low. Many participants traveled significant distances from their remote villages -- often accessible only by foot or basic transportation -- to reach locations with adequate network coverage.

\section{Related Work}

Previous research has mainly examined the intersection between multimodality and cultural diversity, or between multilingualism and cultural knowledge. However, recent efforts indicate a growing shift toward unifying these perspectives----exploring models and benchmarks that simultaneously span multilingual, multimodal, and multicultural dimensions. Benchmarks such as CVQA \cite{romero2024cvqaculturallydiversemultilingualvisual}, ViMUL-Bench \cite{shafique2025culturallydiversemultilingualmultimodalvideo}, ALM-Bench \cite{vayani2025languagesmatterevaluatinglmms}, M5-VGR and M5-VLOD \cite{schneider2024m5diversebenchmark} have been designed to evaluate the performance of models across modalities, languages, and cultures.

\begin{table*}[t]
\centering
\begin{tabular}{lrrrrrS[table-format=6.0]r}
\toprule
\textbf{Language} & \textbf{Recipes} & \textbf{Words} & \textbf{Vocabulary} & \textbf{Recordings} & \textbf{Total Duration (s)} & \textbf{Images} \\
\midrule
Bodo & 95 & 26745 & 3958 & 3043 & 34076 & 1532 \\
Assamese & 113 & 29648 & 757 & 1888 & 7034 & 1415 \\
Meitei & 100 & 18277 & 3474 & 2133 & 16808 & 580 \\
Kaman-Mishmi & 128 & 21334 & 3159 & 3036 & 27207 & 1129 \\
Khortha & 126 & 24398 & 656 & 3131 & 14733 & 1129 \\
Santhali & 120 & 20878 & 380 & 2892 & 12116 & 1004 \\
Ho & 91 & 13767 & 395 & 2300 & 6726 & 875 \\
Sadri & 107 & 13257 & 528 & 2785 & 9139 & 1103 \\
Mundari & 82 & 15243 & 1785 & 2378 & 13432 & 703 \\
Khasi & 98 & 30166 & 1194 & 4534 & 24460 & 1928 \\
\midrule
\textbf{Total} & \textbf{1060} & \textbf{213713} & \textbf{16286} & \textbf{28120} & \textbf{165731} & \textbf{11398} \\
\bottomrule
\end{tabular}
\caption{Recipe Dataset Statistics by Language (Duration in Seconds)}
\label{tab:datastat}
\end{table*}

While efforts to benchmark cultural knowledge and low-resource or endangered languages have evolved independently, especially in a culturally and linguistically diverse country like India, they rarely intersect. For instance, works like SANSKRITI \cite{maji2025sanskriticomprehensivebenchmarkevaluating} and DOSA \cite{seth2024dosadatasetsocialartifacts} focus on evaluating LLMs' understanding of Indian culture and cultural artifacts, whereas efforts like PARIKSHA \cite{watts2024parikshalargescaleinvestigationhumanllm}, INDICGENBENCH \cite{singh2024indicgenbenchmultilingualbenchmarkevaluate}, Indic-QA \cite{singh2025indicqabenchmarkmultilingual}, and MILU \cite{verma2025milumultitaskindiclanguage} provide multilingual benchmarks for Indic language understanding. However, these works remain largely confined to high-resource languages, leaving endangered Indic languages unrepresented. Therefore, in our work, we introduce a cultural knowledge database covering 10 endangered languages of Eastern India. This enables us to evaluate the performance of LLMs on low-resource, culturally grounded tasks, and to highlight the challenges and opportunities in extending language technologies to underrepresented linguistic communities.

% Food as a cultural lens has emerged as a powerful way to capture procedural, linguistic, and socio-cultural narratives with benchmarks like FoodieQA \cite{li2024foodieqamultimodaldatasetfinegrained} and WorldCuisines \cite{winata2025worldcuisinesmassivescalebenchmarkmultilingual} exploring cuisine-centered reasoning, while works on CulturalRecipes \cite{cao2023culturaladaptationrecipes} and CARROT \cite{hu-etal-2024-bridging} explore cultural adaptation of recipes between cuisines (e.g., Chinese and Western) and cross-cultural recipe retrieval. 
% Food, as a cultural lens, has emerged as a powerful medium for capturing procedural, linguistic, and socio-cultural narratives. 

Our work most closely relates to the following: FoodieQA \cite{li2024foodieqamultimodaldatasetfinegrained}, WorldCuisines \cite{winata2025worldcuisinesmassivescalebenchmarkmultilingual}, CulturalRecipes \cite{cao2023culturaladaptationrecipes} and CARROT \cite{hu-etal-2024-bridging}. They focus on cuisine-centered reasoning and cross-cultural recipe adaptation and retrieval. In the Indian context, IndiFoodVQA \cite{agarwal-etal-2024-indifoodvqa} covers several aspects of Indian cuisine and culinary diversity. Even though IndiFoodVQA \cite{agarwal-etal-2024-indifoodvqa} introduces culturally relevant visual question answering grounded in Indian food, it---like other global food datasets---largely targets high-resource settings and fails to represent endangered or minoritized language communities. 

% Additionally, shared tasks such as WMT 2023 \cite{pal-etal-2023-findings} and WMT 2024 \cite{pakray-etal-2024-findings} Shared Task on Low-Resource Indic Language Translation have highlighted progress in low-resource language translation, including some North-Eastern languages, but remain primarily English-centric and fail to capture the deeper cultural context inherent to these languages.

Recipes not only encode procedural knowledge but also serve as inter-generational vessels for transmitting language, values, and identity. In many Eastern Indic communities, making and sharing indigenous recipes sustains linguistic practices endangered in formal or educational settings -- making them a rich and practical source for cultural benchmarking. Hence, in this work, we propose a multilingual benchmark, Endangered Language Recipes (ELR-1000), for Endangered Eastern Languages through Indigenous Recipes that simultaneously targets cultural competence and endangered language preservation.

\begin{figure}[t]
    \centering
    \includegraphics[width=\columnwidth]{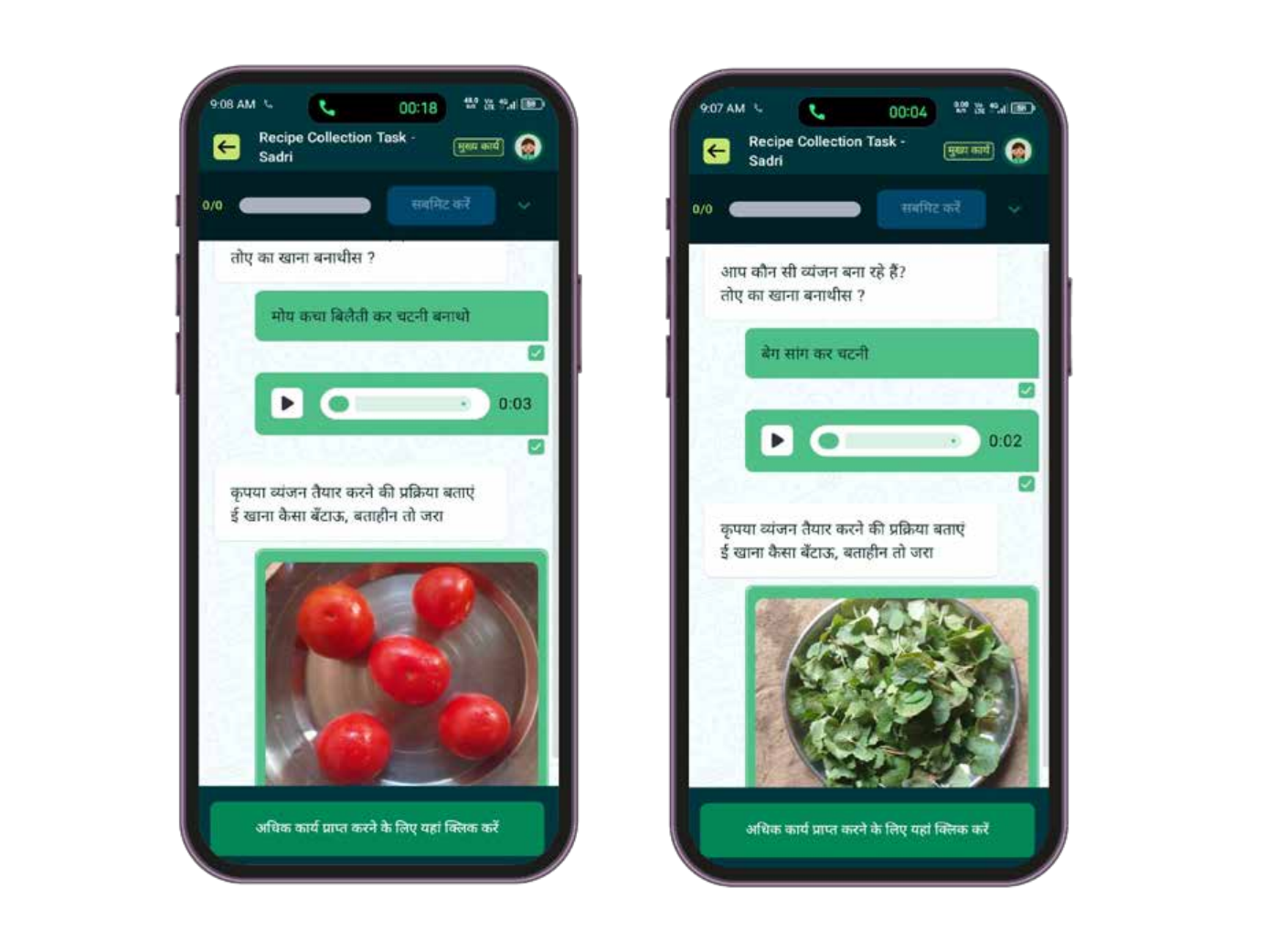}
    \caption{App Interface of the Recipe Documentation Task}
    \label{fig:app_interface}
\end{figure}

% \begin{figure*}[t]
%     \centering
%     \includegraphics[width=0.6\textwidth]{charts/App_interface.png}
%     \caption{Comprehensive Metadata Comparison across Languages}
%     \label{fig:metadata}
% \end{figure*}

% \begin{figure*}[htp]
%     \centering
%     \includegraphics[width=0.8\textwidth]{charts/all_languages_radar_chart.pdf}
%     \caption{Comprehensive Metadata Comparison across Languages}
%     \label{fig:metadata}
% \end{figure*}

\section{Endangered Language Recipes (ELR-1000)}

\subsection{Pilot Study}

The pilot study served to validate our research approach and methodology through engagement with rural women from the Sadri-speaking community. Sadri, classified as an under-resourced language, is spoken by approximately 5.1 million people primarily in the eastern regions of India. This linguistic community was strategically selected for our pilot experiments because of Karya's established relationships within these communities and familiarity with the Sadri language, which facilitated authentic engagement and reduced potential barriers to participation. After receiving a positive response from the participants, we organized a demonstration session with approximately 30 Sadri-speaking women from 3 remote tribal villages in Jharkhand. In this session, each participant recorded one recipe using the Karya application. 
% After receiving a positive feedback from the participants, Karya organized a demonstration session with approximately 20 Sadri-speaking women from remote tribal villages, where each participant recorded one recipe using the Karya application.
Notably, all of these women were new to digital work and represented communities with limited access to technology and digital literacy programs. The positive response and successful completion of this pilot study motivated us to expand the project.
% The enthusiastic response and successful completion of this pilot study convinced us to expand the project.
% Based on the pilot results, we selected 10 endangered languages as per the UNESCO Endangered Languages list\footnote{\url{https://unesdoc.unesco.org/ark:/48223/pf0000192416.locale=en}}, with 5 from Jharkhand and Bihar, and 5 from Northeast India, balancing high-resource and low-resource languages to ensure comprehensive representation. 
\subsection{Language Selection}
Based on the successful completion of the pilot project, we selected ten endangered languages from the UNESCO Endangered Languages list.\footnote{\url{https://unesdoc.unesco.org/ark:/48223/pf0000192416.locale=en}} Of these, five are majorly spoken in Jharkhand and Bihar, and the rest are from Northeast India. We aimed to balance high-resource and low-resource languages to ensure a representative sampling.
These languages are characterized by their geographic isolation, with speakers concentrated in remote tribal areas having negligible digital presence or technological resources. The selected languages allowed for regional and linguistic diversity. Thereby, the resulting dataset reflected variations in cooking techniques, local crop varieties, ingredient naming conventions, and cultural traditions that differ significantly across India’s ecological zones---offering an inclusive and representative portrait of the country's rich food heritage. Table \ref{tab:datastat} shows the collected dataset statistics.

The selected languages were:
\begin{itemize}
    \item \textbf{Languages from Jharkhand and Bihar}: Ho (spoken by approximately 1.04 million people in remote tribal areas), Khortha (limited to specific districts with minimal written documentation), Sadri (scattered across isolated mining regions), Santhali (primarily in rural tribal communities), and Mundari (concentrated in remote forest areas with limited connectivity)
    \item \textbf{Languages from North Eastern region of India}: Assamese, Meitei, Khasi, Bodo (concentrated in isolated areas of Assam), and Kaman Mishmi (critically endangered with speakers in extremely remote border regions)
\end{itemize}

% Our project successfully collected approximately 1,060 traditional recipes. All of these were contributed by tribal and non-tribal women from some of the most geographically isolated and technologically underserved regions of India. 

% Our project successfully collected approximately 1,060 traditional recipes, all contributed by tribal and non-tribal women from some of the most geographically isolated and technologically underserved regions of India. These contributors come from communities where internet connectivity is sporadic, smartphone penetration is minimal, and digital literacy rates remain exceptionally low. Many participants traveled significant distances from their remote villages -- often accessible only by foot or basic transportation -- to reach locations with adequate network coverage.

We primarily aimed at women as participants in this study. During the mobilization process, we selected women over men based on the local employment context: women in these villages have significantly fewer income-generating opportunities due to household responsibilities that typically keep them at home. This made them well-suited for our data collection,they had the flexibility to participate, could complete tasks from their homes without needing to travel, and possessed deep generational knowledge of traditional culinary practices developed through years of cooking for their families. This approach enabled us to efficiently gather rich culinary data while simultaneously providing income opportunities to a demographic with limited access to formal employment.

% Each recipe is documented in a structured, multimodal format including text, audio, and images, captured in their native languages. 

%This dataset is unique in that it documents authentic and underrepresented culinary knowledge from some of India’s most marginalized linguistic communities. It encodes this knowledge in a machine-interpretable format, enabling computational access and analysis. In doing so, it preserves not only traditional recipes, but also the cultural narratives, cooking techniques, and indigenous food wisdom they embody -- forms of knowledge that are at risk of permanent loss
% The uniqueness of this dataset lies in the fact that it captures authentic, underrepresented culinary knowledge from India's most marginalized linguistic communities in a machine-interpretable form, preserving not just recipes but also cultural narratives, traditional cooking techniques, and indigenous food wisdom that might otherwise be lost forever.

% \subsection{Regional \& Linguistic Diversity}
% To ensure broad cultural and ecological coverage, we selected 10 endangered languages spanning Jharkhand, Bihar, and Northeast India. These include Ho, Khasi, Khortha, Sadri, Santhali, Mundari, Assamese, Meitei, Bodo, and Kaman Mishmi. This regional and linguistic diversity allows the dataset to reflect variations in cooking techniques, local crop varieties, ingredient naming conventions, and cultural traditions that differ dramatically across India’s ecological zones -- offering an inclusive and representative portrait of the country’s rich food heritage.

\subsection{Application Design \& Usability Considerations}
The Karya mobile application was designed for rural, often first-time digital workers with limited education or smartphone experience. To support varied literacy levels, the app used a minimal-text interface with clear audio-visual cues in local languages.
% Recognizing the diverse literacy levels among participants, the application prioritized ease of use through a minimal-text interface enhanced with clear audio-visual cues in local languages. 
Instructions for each recording step were delivered via simple icons and audio guidance to reduce cognitive load and accommodate non-literate users.

Importantly, the design allowed participants to review and edit their text or audio entries before final submission allowing them to correct mistakes independently. To avoid interrupting cooking, the application separated media capture from annotation: users first took photos, then added text or audio explanations later at their own pace. To handle poor connectivity in remote areas, the application also supported offline data entry, letting users save sessions locally and upload them when internet was available. Figure \ref{fig:app_interface} shows Karya application's interface.

% As recording detailed instructions while actively cooking would be difficult, the app workflow separated media capture from annotation. Users could first take a series of photos documenting the cooking process without interruption. Later they can add text or audio explanations to each image at their own pace.

% Another critical feature addressed connectivity challenges common in remote areas. The application was fully offline-enabled for data entry, allowing contributors to save their complete recording sessions locally on their devices. Submissions could be uploaded to the server as and when internet connectivity became available, ensuring that even participants in areas with patchy or unreliable networks could contribute without technical barriers.

\subsection{Capacity Building Trainings}

To ensure effective community engagement, we recruited one local coordinator per language---each a native speaker residing in their respective area---with the help of local Non-Governmental Organizations (NGOs) working in each region. These coordinators then mobilized 30-50 rural women participants per language (aged 15-45), targeting individuals who were fluent in reading, writing, and speaking in their native language and had smartphone access.

After recruitment, Karya held in-person training and application demos to guide coordinators and participants on effective recording. We also explained the project’s goal of preserving traditional recipes in native languages. The local coordinators played a crucial role in bridging the gap between us and the community members, by explaining tasks to participants, and providing ongoing support when people faced challenges with recording in the application. Participants were asked to provide their informed consent before starting the data collection tasks.

% Once participants were recruited, [REDACTED] conducted physical training sessions and application demonstrations in these areas to ensure coordinators and participants understood how to record the steps effectively in the application. We also explained the project objectives to participants, emphasizing the importance of collecting traditional recipes in their native language for cultural preservation purposes.

% The local coordinators played a crucial role in bridging the gap between us and the community members, by explaining tasks to participants, and providing ongoing support when people faced challenges with recording in the application. 

% Given that many participants came from rural areas with limited digital literacy, we designed a user-friendly three-phase data collection process.

Each recipe submission underwent a thorough review and validation by the local coordinator. Since many contributors were first-time digital workers, this validation process played a vital role in ensuring accuracy and quality. Once approved, contributors received transparent payments directly through the Karya application, earning \$8.68 per recipe. On average, participants submitted between two to five recipes, allowing them to earn from \$17.50 up to \$43.50 for their valuable contributions.

%\subsection{Validation \& Payment Process}
%Once each recipe was recorded, the local coordinator acted as a validator, reviewing and validating the recipes at the participant level. After the coordinator approved the submissions, participants were prompted to enter their bank account or UPI details into the Karya app wallet system. The application displayed transparent wage information, showing each participant's earnings based on their completed recipes.

%To ensure secure payment processing, we implemented a two-step verification system. Once participants entered their banking details, they received a test payment of \$0.023. Upon confirmation of receiving this test amount, participants were given full payment within one week. This validation and payment process ensured both quality control and secure, transparent compensation for all participants.

%\subsection{Remuneration to participants} 
%Remuneration rates were fixed at \$8.68 for every recipe that a participant provided. Each participant contributed 2 to 5 traditional recipes using the Karya mobile application, earning between \$17.5 and \$43.5. They received  for each recipe they submitted as part of the recipe collection initiative. %The task was designed to be simple and accessible, enabling women to easily record their contributions. The process involved taking photographs, recording audio, and adding text annotations.

\subsection{Addressing Multimodal Data Heterogeneity}

The data collection framework prioritized contributor authenticity over structural consistency, allowing participants the freedom to document culinary practices in their own style using the Karya application. This intentional flexibility captured rich cultural and epistemic representations of food knowledge but introduced substantial structural heterogeneity across the dataset. Table \ref{tab:datastat} compares the metadata across all languages.

Participants used different strategies to contribute---some provided detailed text with images, others relied heavily on audio narration--- resulting in variable completeness across modalities. This created a complex, un-standardized dataset requiring systematic normalization.

To handle this, we implemented a modular, array-based structure separating text, image, and audio content. Each recipe step was decomposed and stored in parallel directories, with pre-processing steps to address null values and corrupted media. Quantitative analysis revealed diverse documentation patterns: 54\% of recipe steps included all three modalities, 29\% used image-text, and 83.4\% had featured text. Notably, 64.5\% included audio narration, underscoring the cultural significance of oral knowledge. These patterns demonstrate that when unconstrained, contributors favored multimodal formats that reflect oral traditions and the limitations of purely textual expression.

\section{Challenges Faced During the Collection of ELR-1000}
A key contribution of this project lies in documenting data collection challenges. And this can inform future efforts. These challenges fall into the following broad categories:

% A significant contribution of this project is in the challenges faced during data collection because these experiences can contribute towards building a template for similar data collection initiatives. The challenges faced can be broadly categorized into the following:

\paragraph{Demographic \& Socio-Cultural Challenges} Digital literacy among participants varied widely, with many experiencing structured documentation and in-app recording for the first time. This called for intensive, culturally sensitive training to ensure comfort and understanding.
% Participants were primarily rural women, many of whom were first-time users of smartphones and digital applications. 
% Digital literacy varied significantly even within a single community. For many, this was their first experience with structured documentation work, or with in-app recording of text, images, and audio. This required intensive, culturally sensitive training sessions to ensure comfort and comprehension.

\paragraph{Trust building} Some communities were initially skeptical about sharing traditional knowledge and our project's goals. To build trust, local coordinators explained the project in familiar terms, and we maintained transparency around payments, data use, and cultural preservation goals.

% In some regions, there was initial skepticism about the intentions behind the project, especially around sharing traditional knowledge with outsiders or fears that their knowledge might be exploited without fair compensation. To address this, we worked closely with local and respected community coordinators, who explained the project goals in local languages and contexts. Transparency about payment, usage of data, and cultural preservation aims helped foster trust over time.

\paragraph{Seasonality of Food Practices} A key challenge in documenting tribal and rural recipes was their seasonality and reliance on foraged ingredients. Some dishes couldn’t be recorded year-round due to ingredient unavailability. For example, some forest greens or wild fruits are harvested only during the monsoon or winter seasons.  To address this, we encouraged participants to note seasonal variants and substitutions in the cultural context section.

% A unique challenge in documenting tribal and rural recipes lies in their deep seasonality and reliance on foraged or locally grown ingredients. Certain dishes could not be recorded during specific months because their key ingredients were unavailable. For example, some forest greens, tubers, or wild fruits are harvested only during the monsoon or winter seasons. This introduced variability in recipe availability, requiring us to accept that our initial dataset would necessarily be partial and context-specific. We addressed this by encouraging participants to note seasonal variants and ingredient substitutions during the cultural context section of the recording.

% \paragraph{Infrastructure \& Connectivity Limitations} Many participants lived in areas with unreliable internet connectivity and power supply. Uploading multimodal data (images, audio) from remote villages posed logistical difficulties. To mitigate this, the app was designed to support offline recording with deferred upload when connectivity improved. Coordinators also helped participants schedule recording sessions in areas with better signal coverage or transported recordings to locations with stronger networks.

\paragraph{Ensuring Fair Payment} A nuanced challenge was ensuring fair compensation across recipes of varying complexity. While some involved extensive effort and ancestral techniques, others were simple but equally authentic. Paying a fixed rate felt unfair to those doing more work, yet dismissing simple dishes was not culturally appropriate. We addressed this by setting a minimum step requirement to ensure baseline documentation quality.

% One nuanced challenge was determining fair compensation for recipes that varied greatly in complexity and effort. Some participants contributed elaborate dishes that required foraging ingredients from forests at dawn and cooking for hours using complex, ancestral techniques. Others recorded simpler recipes like boiled vegetables. Both are equally authentic and culturally important but one involves fewer steps. We recognized that culturally, it would be inappropriate to dismiss such simple dishes as “less valuable”. However, paying the same fixed rate for all recipes sometimes felt inequitable to those putting in greater effort. To address this, we introduced a minimum requirement for the number of preparation steps, ensuring a baseline of documentation quality.

\paragraph{Maintaining Data Quality \& Consistency} Given the diversity of languages, cooking styles, and literacy levels, ensuring consistency and quality in the recordings was another major challenge. Recipe steps varied in granularity, and the local names of ingredients often lacked standardized spellings. To manage this, we implemented a two-layer validation system: local coordinators first reviewed entries for completeness and clarity, and project managers conducted spot checks across languages for cross-regional consistency.

\paragraph{Resolution through Iterative Feedback \& Capacity Building} One-time training was not sufficient, so we adopted an iterative support model. Coordinators regularly assisted participants, resolved issues, and encouraged re-recordings -- crucial for maintaining motivation and improving quality.

\section{Experimental Design and Methodology}
\paragraph{Research Motivation} The primary motivation for this research was to assess the current capabilities of large language models (LLMs) regarding cultural relevance in the food or recipe domain, specifically for communities that speak and write under resourced East Indic languages.
To this end, the experiment was designed to evaluate LLM translations of traditional recipes into English. The evaluation specifically focused on the models' ability to maintain cultural authenticity and factual correctness, process local dialects, and preserve the original instructional format of the recipes.
\subsection{Experimental Framework} To structure our experiments, we divided the evaluation based on the capabilities of different machine learning models. Neural Machine Translation (NMT) models typically have sentence-level context windows, while LLMs possess much longer context windows that enable them to understand recipe structures and contextual information within the prompt, even when direct translation capabilities may be limited.

\subsubsection{Neural Machine Translation (NMT) Evaluation} The majority of languages in our dataset are not supported by state-of-the-art NMT systems. Since Assamese is a relatively resource-rich language compared to others in our study, many leading NMT systems did provide support for it. However, both BLEU and chrF scores were significantly lower than expected. We attribute to the specialized domain of traditional recipes and the limited training data for such culturally specific content.
An additional challenge emerged from script variations. While NMT models typically support only Devanagari script for certain languages, some participants in our data collection process provided recipes in Latin script. This occurred because we prioritized authentic cultural expression over standardized orthography, allowing participants complete freedom in their creative expression to capture genuine traditional recipes.

\subsection{Large Language Model Evaluation}

\subsubsection{Model Selection and Setup}
Since the primary objective of this study was to evaluate translation quality, model selection focused on their documented strengths in handling multilingual inputs, contextual understanding, and cultural relevance in translation outputs. To ensure a balanced evaluation, we selected an equal number of proprietary and open-source models three proprietary models (Gemini 2.5 Flash, GPT-4o, and Claude Sonnet 4) and three open-source models (Llama 4 Scout 17B-16E, Mistral Small 3.1 (25.03), and CohereLabs Aya Expanse 8B). This approach allowed us to maintain fairness in comparing models across different licensing types, avoiding any inherent bias towards either commercial or open alternatives.
Additionally, we prioritized models that represent the best publicly accessible options in their respective categories. Care was taken to avoid comparisons between models of vastly different capacities (e.g., large reasoning models versus small open models) to ensure that the evaluation remains meaningful, relevant, and reflective of practical use cases.
We evaluated six state-of-the-art LLMs using Gemini 2.5 Flash (on default settings) as the evaluation judge. LLMs possess longer context windows and often contain cultural and demographic information about endangered Indic languages within their training data. However, most top-performing models do not possess direct capabilities to translate these specific languages into English.

\subsubsection{Experimental Conditions}
We conducted two experiments for each LLM:
\paragraph{No Context Condition} We assessed baseline model performance by providing only the complete recipe in the source language and requesting translation to English. This condition evaluated the models' inherent knowledge and translation capabilities without additional guidance.

\paragraph{Contextual Condition} We enhanced the translation context by providing four things: (1) Background information about the source language and the communities which speak them (2) Few-shot translation examples. (3) Specific guidelines for cultural preservation in translation (4) Instructions for maintaining recipe structure and terminology.
    % \begin{itemize}
    %     \item Background information about the source language and the communities which speak them.
    %     \item Few-shot translation examples.
    %     \item Specific guidelines for cultural preservation in translation.
    %     \item Instructions for maintaining recipe structure and terminology.
    % \end{itemize}

The results showed significantly improved performance across all models in the contextual condition, with particularly dramatic improvements for models like Mistral that exhibited hallucination behaviors without context.

\subsection{Evaluation Methodology}

\subsubsection{Human-in-the-Loop Evaluation Protocol}
Our evaluation followed a rigorous three-stage process and the metrics used are in Appendix \ref{sec:evalm}:
\begin{itemize}
    \item \textbf{Human Translation:} We first obtained high-quality human translations for all test recipes from native speakers of each language.
    
    \item \textbf{LLM Translation Generation:} We processed the raw recipe data through each LLM under both experimental conditions (no context and contextual).
    
\item \textbf{Hybrid Evaluation:} We employed a two-judge LLM ensemble, Gemini 2.5 Pro and OpenAI GPT-5, for automated evaluation against human reference translations. This ensemble was chosen to enhance the robustness and reliability of the scoring by mitigating model-specific biases inherent in single-judge systems. Following the automated scoring, Human Oversight was applied to a sample of evaluations, particularly those with score discrepancies, to verify and validate the results and ensure ultimate accuracy.
\end{itemize}

\begin{figure}[t]
    \centering
    \includegraphics[width=0.45\textwidth]{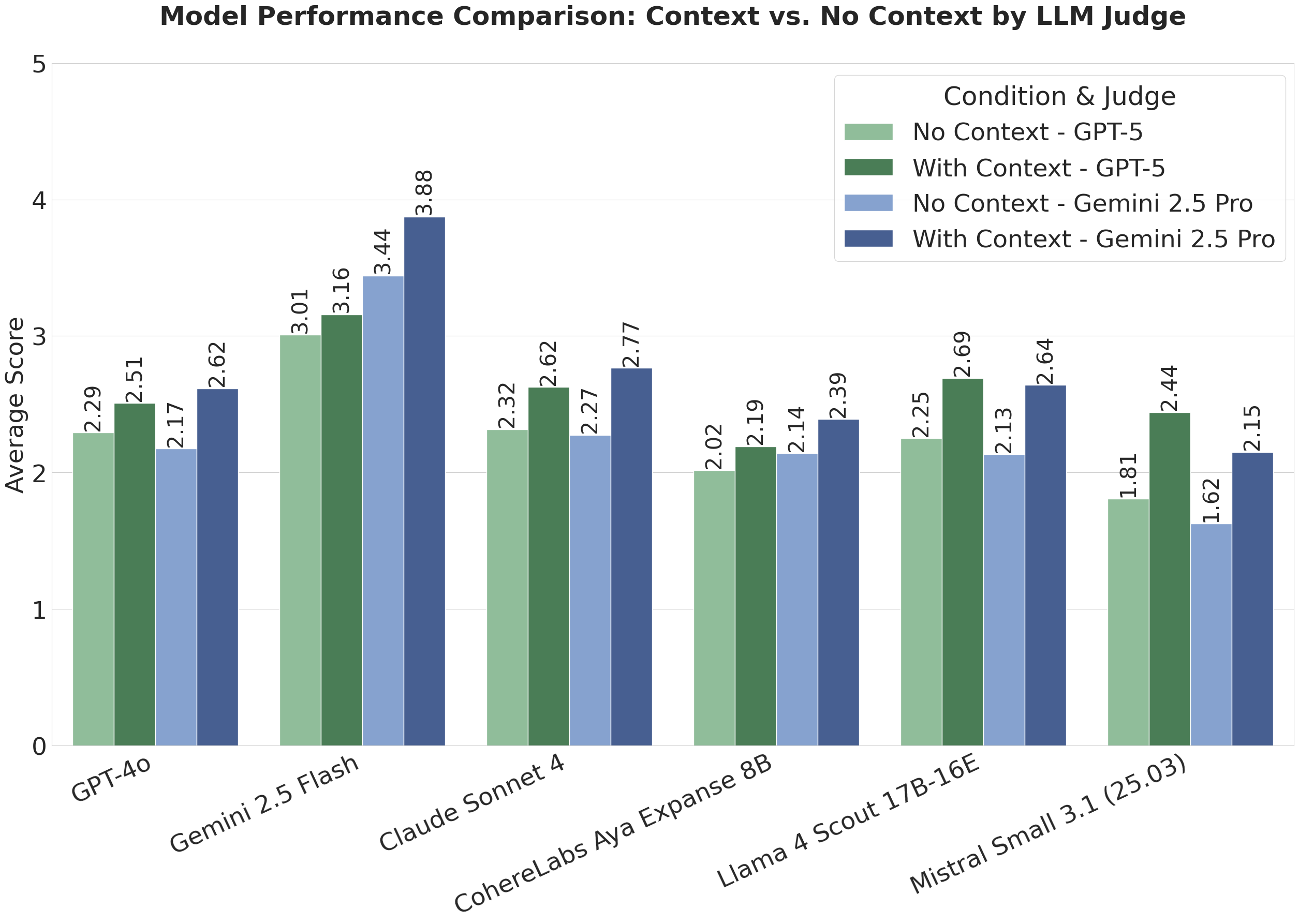
}
    \caption{Impact of Contextual Information on Model Performance}
    \label{fig:context}
\end{figure}
\subsection{Results}

\subsubsection{Overall Model Performance}
Analysis across all ten languages reveals distinct performance tiers among the evaluated models. The provision of context was universally critical, dramatically separating useful translations from unusable ones.

\subsubsection{Top Performers}
Gemini 2.5 Flash consistently emerged as the most capable and reliable model, achieving the highest average scores in both experimental conditions. In the contextual setting, it frequently received perfect or near-perfect ratings, demonstrating exceptional strength in handling complex cultural translations across languages such as Santhali, Meitei, and Assamese.

Claude Sonnet 4, GPT 4o and Llama-4-Scout-17B formed the second performance tier in contextual conditions. Both models showed significant improvement with context, with Llama-4-Scout demonstrating particular strength in Sadri and Ho translations, while Claude Sonnet 4 excelled in Khortha and Ho recipe translations.

\subsubsection{General Performance Patterns}
Without context, most models struggled significantly, with average scores indicating that translations were largely unusable for practical applications. As shown in Table \ref{tab:translation-comparison}, models like Gemini frequently produced generic or misplaced substitutions—such as replacing star fruit with bamboo shoot—and omitted culturally specific implements like the mortar and pestle. When contextual information was provided, performance improved across all models: translations became more faithful to the source and better reflected local tools and ingredients.

However, even with context, human translations remained superior, particularly in preserving cultural nuance. For instance, Gemini introduced terms such as chopping board in both contextual and non-contextual outputs—a tool almost never used in indigenous kitchens—revealing a persistent bias toward Western culinary concepts. This highlights that while contextual information aids fluency and accuracy, current LLMs still tend to normalize outputs toward globalized or Western frames of reference, underscoring the need for culturally grounded datasets and evaluation frameworks.

\begin{table*}[t]
\centering
\small
\setlength{\tabcolsep}{6pt} % Adjusts horizontal padding
\renewcommand{\arraystretch}{1.2} % Slightly increases row height

\begin{tabular}{p{4cm}p{5cm}p{6cm}}
\toprule
\textbf{Source} & \textbf{Sentence 1} & \textbf{Sentence 2} \\
\midrule
\textbf{Human Translation} &
I’m going to cook star fruit and fish curry. &
To cook this I have used a wok, a lid, a ladle, mortar and pestle, a bowl, knife, and a vegetable basket. \\
\addlinespace[0.3em]
\textbf{Gemini (No Context)} &
I am going to cook a dish with \textbf{bamboo shoot and fish}. &
To cook this, I used a pan, knife, \textbf{cutting board}, spatula, bowl, ladle, and a gas stove. \\
\addlinespace[0.3em]
\textbf{Gemini (With Context)} &
I am going to cook \textbf{star fruit with fish} now. &
For cooking this, I have used a mortar and pestle, \textbf{chopping board}, knife, stirring spoon, large cooking pot, cleaver, and a bowl. \\
\bottomrule
\end{tabular}

\caption{Comparison between Human Translation and Gemini outputs (with and without context).}
\label{tab:translation-comparison}
\end{table*}

\section{Discussion \& Analysis}

\subsection{The Paradox of Fluent Falsehood}
The most salient observation from our analysis reveals a critical disconnect between translation fluency and content accuracy. Across nearly all models and languages, ``Fluency'' and ``Comprehensibility'' scores consistently exceeded ``Adequacy'' and ``Cultural Appropriateness'' scores. This pattern indicates that models excel at generating grammatically correct, readable English output that often bears minimal resemblance to the source text's actual meaning.

This phenomenon, which we term ``fluent falsehood'', creates a dangerous illusion of successful translation. For the most challenging languages such as Bodo and Kaman Mishmi, models did not simply mistranslate; they systematically hallucinated entirely different recipes. Traditional ingredients like silkworms or specific regional vegetables were replaced with generic instructions for mushrooms, chicken curry, or completely unrelated dishes.

\subsection{Context as a Critical Success Factor}
Context provision emerged as the single most important factor determining translation success or failure. For models like Mistral, contextual information represented the difference between nonsensical output (scoring 1.0) and usable translations (scoring 4.0 or higher).

Context proved especially vital for cultural appropriateness. Models without context consistently failed to translate traditional tool names, culturally specific cooking practices, or unique ingredients, instead opting to omit them or substitute generic Western equivalents. Gemini 2.5 Flash's cultural handling in Santhali and Meitei translations improved to near perfect scores when provided with appropriate context in fluency and comprehensibility metrics. Figure \ref{fig:context} shows how context improves different LLMs' performance on our benchmark. 

\subsection{Cultural Blindness and Systematic Errors}
Our analysis revealed profound model limitations in understanding and preserving cultural and material context:
\begin{itemize}
    \item \textbf{Systematic Ingredient Misidentification:} Models repeatedly misidentified key recipe components. ``Jhingi'' was consistently mistranslated as ``prawns/shrimp'', while Bodo silkworm recipes were altered to feature mushrooms or conventional meat. These errors fundamentally alter the dish's cultural and nutritional identity.
    
    \item \textbf{Loss of Traditional Methods:} Cultural nuances related to traditional cooking methods, specialized utensils, and embedded cultural narratives were lost in translation, particularly without contextual guidance.
\end{itemize}

\subsection{Language-Specific Performance Variations}
Model performance varied noticeably across language families. Translation quality was notably better for languages such as Khortha and Sadri, while languages including Kaman Mishmi, Bodo, Ho, Santhali and Mundari prompted near-complete translation failures across most models. This variation suggests that latent linguistic knowledge from pre-training data may provide differential support for certain languages, even within the ``low-resource'' category.

\subsection{Implications for Endangered Language Documentation}
This comprehensive analysis reveals that while LLMs possess sophisticated text generation capabilities, their application to endangered language translation presents significant challenges. The models frequently function as ``fluent fabricators'' rather than faithful translators, emphasizing the critical need for contextual information, evaluation beyond surface-level fluency metrics, and human oversight in endangered language NLP efforts.

% \begin{figure*}[htp]
%     \centering
%     \includegraphics[width=0.8\textwidth]{charts/all_languages_radar_chart.pdf}
%     \caption{Comprehensive Metadata Comparison across Languages}
%     \label{fig:mychart}
% \end{figure*}

% \begin{figure*}[htp]
%     \centering
%     \includegraphics[width=0.8\textwidth]{charts/context_comparison_bar_chart-1.pdf}
%     \caption{Impact of Contextual Information on Model Performance}
%     \label{fig:mychart}
% \end{figure*}
\section{Conclusion}

%This study addresses the need for culturally grounded and linguistically inclusive benchmarks in NLP, particularly for endangered and low-resource languages. By using traditional recipes as a rich, multimodal lens into language and culture, we present a dataset that bridges computational research with community-driven knowledge preservation. Our findings reveal the limitations in current translation models, especially their tendency to generate fluent yet culturally inaccurate outputs in the absence of contextual guidance. These patterns highlight the inadequacy of standard evaluation metrics and stress the importance of human-in-the-loop methodologies and culturally sensitive prompts. As large language models continue to expand their reach, our work offers a foundation for building more equitable, ethical, and context-aware AI systems. We also document the unique challenges faced during data collection and validation as a practical guide for others seeking to build similar community-rooted, multimodal datasets. We see this dataset as a vital intervention in preserving the linguistic and cultural heritage embedded in everyday food practices. It also advances more meaningful evaluation of translation systems for underrepresented languages before such knowledge disappears from both community memory and the digital world.

This study addresses the need for culturally grounded and linguistically inclusive benchmarks in NLP, particularly for endangered and low-resource languages. By using traditional recipes as a rich, multimodal lens into language and culture, we present a dataset that bridges computational research with community-driven knowledge preservation. Our findings reveal the limitations in current translation models, especially their tendency to generate fluent yet culturally inaccurate outputs in the absence of contextual guidance. These patterns highlight the inadequacy of standard evaluation metrics and stress the importance of human-in-the-loop methodologies and culturally sensitive prompts. 

Our findings further suggest that translation errors in low-resource, culturally specific data are not merely linguistic but \textit{epistemic} ---arising from a lack of cultural grounding within current large language models. As observed in ELR-1000, models often replace indigenous tools, practices, or ingredients with globally dominant equivalents, reflecting their limited awareness of cultural context. Future research could therefore move beyond surface-level evaluation and explicitly probe LLMs’ cultural awareness through question–answering tasks or internal probing methods that examine how cultural concepts are represented in model embeddings. Such investigations could reveal how cultural information (or its absence) propagates within translation paradigms, and how integrating culturally grounded datasets like ELR-1000 can enable models to move from syntactic fluency to contextual fidelity, preserving the lived realities embedded in endangered languages.

As large language models continue to expand their reach, our work offers a foundation for building more equitable, ethical, and context-aware AI systems. We also document the unique challenges faced during data collection and validation as a practical guide for others seeking to build similar community-rooted multimodal datasets. We see this dataset as a vital intervention in preserving the linguistic and cultural heritage embedded in everyday food practices. It also advances more meaningful evaluation of translation systems for underrepresented languages before such knowledge disappears from both community memory and the digital world.

\section*{Limitations}
There are several limitations given the fact that this is only the beginning of what we believe should be a much larger community-authored effort. We list some of them below
\begin{itemize}
    \item Translations were not possible for the full dataset in this phase as recruiting bilingual speakers was not the focus of this study---our primary aim was to document knowledge from native speakers, many of whom were monolingual. As a result, only a three recipes per language were translated and used for evaluation. This limits the generalizability of conclusions. 
    % The absence of established benchmarks for these languages further complicates performance validation. 
    % \item We translated only 3 recipes per language to benchmark the LLMs. While getting recipes translated manually took a significant effort from our end, we understand that this number may not be enough for make strong conclusions about the LLMs ability to translate these recipes in endangered languages to English. In our future work, we plan to expand this and translate all the recipes released as part of this work.
    \item In this work, we work on building resources for 10 endangered languages. While we believe this to be a great beginning, especially considering that all of these languages are endangered, we would like to scale our efforts to cover more such languages in the future. We believe that there is a great scope in this direction given the diversity of languages present in Eastern part of India.
    \item We collect approximately 100 recipes per language and release it in ELR-1000. This may be sufficient for benchmarking existing LLMs but may not enough to improve the LLMs to work better in this domain and language. In the future, we would like to work with more rural communities to crowdsource more recipes so that the collected dataset can be used to improve the abilities of LLMs in this domain.
    \item In this project, we limited ourselves to the cuisine domain. We did this so that the collected dataset is culturally relevant to the communities we worked with. However, cuisine is not the only topic that could help us collect culturally relevant data. Some other such topics could be the agricultural and livestock farming practices that these communities follow. Covering many diverse topics like this could make the collected benchmark more valuable. In the future, we would like to expand our efforts in this direction as well.
    \item Finally, although our dataset is multimodal---including audio, text, and images---it is not aligned to support benchmarking LLMs for advanced tasks such as knowledge graph construction or multimodal reasoning.
\end{itemize}

\section*{Acknowledgments}
We gratefully acknowledge the local coordinators and participants whose time, knowledge, and effort made this work possible. We extend our sincere thanks to Divyansh and Abhijeet for providing technical support throughout this study. We are also deeply grateful to Saroj Kumar and Pradan Katoria for their invaluable assistance in identifying and connecting us with local coordinators.

% This document has been adapted
% by Steven Bethard, Ryan Cotterell and Rui Yan
% from the instructions for earlier ACL and NAACL proceedings, including those for
% ACL 2019 by Douwe Kiela and Ivan Vuli\'{c},
% NAACL 2019 by Stephanie Lukin and Alla Roskovskaya,
% ACL 2018 by Shay Cohen, Kevin Gimpel, and Wei Lu,
% NAACL 2018 by Margaret Mitchell and Stephanie Lukin,
% Bib\TeX{} suggestions for (NA)ACL 2017/2018 from Jason Eisner,
% ACL 2017 by Dan Gildea and Min-Yen Kan,
% NAACL 2017 by Margaret Mitchell,
% ACL 2012 by Maggie Li and Michael White,
% ACL 2010 by Jing-Shin Chang and Philipp Koehn,
% ACL 2008 by Johanna D. Moore, Simone Teufel, James Allan, and Sadaoki Furui,
% ACL 2005 by Hwee Tou Ng and Kemal Oflazer,
% ACL 2002 by Eugene Charniak and Dekang Lin,
% and earlier ACL and EACL formats written by several people, including
% John Chen, Henry S. Thompson and Donald Walker.
% Additional elements were taken from the formatting instructions of the \emph{International Joint Conference on Artificial Intelligence} and the \emph{Conference on Computer Vision and Pattern Recognition}.

\bibliography{custom}
\onecolumn
\appendix

\section{Evaluation Metrics}
\label{sec:evalm}
We employed a comprehensive 5-point Likert scale evaluation across four critical dimensions:

\begin{enumerate}
    \item \textbf{Adequacy (Meaning Preservation):}
    \begin{itemize}
        \item 1: No meaning preserved
        \item 2: Minimal meaning preserved
        \item 3: Moderate meaning preserved
        \item 4: Mostly preserved with minor omissions
        \item 5: Fully preserved meaning
    \end{itemize}
    
    \item \textbf{Fluency (Grammatical Correctness):}
    \begin{itemize}
        \item 1: Extremely poor fluency
        \item 2: Poor fluency with multiple errors
        \item 3: Acceptable fluency, some errors
        \item 4: Good fluency, few errors
        \item 5: Excellent fluency, no errors
    \end{itemize}
    
    \item \textbf{Comprehensibility (Target Language Understanding):}
    \begin{itemize}
        \item 1: Completely incomprehensible
        \item 2: Difficult to understand
        \item 3: Understandable with effort
        \item 4: Mostly comprehensible
        \item 5: Fully comprehensible
    \end{itemize}
    
    \item \textbf{Cultural and Contextual Appropriateness:}
    \begin{itemize}
        \item 1: Inappropriate or offensive
        \item 2: Significant cultural inaccuracies
        \item 3: Some inaccuracies but generally acceptable
        \item 4: Mostly appropriate with minor issues
        \item 5: Fully appropriate and well-adapted
    \end{itemize}
\end{enumerate}

Each translation pair received scores across all four metrics, with human evaluators providing justifications for scores to ensure consistency and reliability in the evaluation process.

% --- APPENDIX SECTION 1: PROMPTS ---
\section{Prompts}
\label{sec:prompts}
This section contains the system and generation prompts used in our experiments and the evaluation prompts for the LLM Judge.

\subsection{Context Free System Prompt}
\begin{promptbox}
You are an expert language translator. Please translate the given text from \{lang\} to English.
\end{promptbox}

\subsection{Context Aware System Prompt}

\begin{promptbox}
You are a specialized linguist and cultural translator with expertise in endangered languages that have minimal digital documentation. Your mission is to provide accurate, culturally sensitive translations from \{lang\} to English while preserving the linguistic and cultural integrity of the source material.

\subsection*{Language Background: Understanding \{lang\}}

\subsubsection*{Geographic and Cultural Context}
\{lang\} (also known as Nagpuri or Kurukh Sadri) is an Indo-Aryan language primarily spoken in:
\begin{itemize}
    \item \textbf{Jharkhand} (main concentration in Ranchi, Gumla, Simdega districts)
    \item \textbf{West Bengal} (Purulia district)
    \item \textbf{Odisha} (Sundargarh district)
    \item \textbf{Assam} (tea garden communities)
\end{itemize}

\subsubsection*{Speaker Communities and Cultural Significance}
\begin{itemize}
    \item \textbf{Primary Speakers}: Kurukh/Oraon tribal communities, Munda speakers, and other Adivasi groups
    \item \textbf{Total Speakers}: Approximately 2-3 million (declining)
    \item \textbf{Cultural Role}: Language of inter-tribal communication, traditional storytelling, folk songs, and cultural ceremonies
    \item \textbf{Social Context}: Often used as a lingua franca among different tribal communities in the region
\end{itemize}

\subsubsection*{Linguistic Characteristics Affecting Translation}
\paragraph{Script and Writing System}
\begin{itemize}
    \item \textbf{Traditional}: Devanagari script (as seen in examples)
    \item \textbf{Status}: Limited standardized orthography; oral tradition predominant
    \item \textbf{Challenge}: Spelling variations common due to lack of standardization
\end{itemize}

\paragraph{Key Grammatical Features}
\begin{enumerate}
    \item \textbf{Word Order}: Subject-Object-Verb (SOV) structure
    \item \textbf{Agglutination}: Suffixes attached to root words for grammatical meaning
    \item \textbf{Case System}: Nominative, accusative, genitive, and locative markers
    \item \textbf{Verb Conjugation}: Complex tense-aspect system with evidentiality markers
    \item \textbf{Honorific System}: Respectful and familiar speech levels
\end{enumerate}

\paragraph{Vocabulary Characteristics}
\begin{itemize}
    \item \textbf{Core Vocabulary}: Mix of Indo-Aryan base with significant tribal language borrowings
    \item \textbf{Cultural Terms}: Rich vocabulary for:
    \begin{itemize}
        \item Traditional foods and cooking methods
        \item Forest products and gathering practices
        \item Agricultural terms and seasonal activities
        \item Kinship and social relationships
        \item Religious and ceremonial concepts
    \end{itemize}
    \item \textbf{Code-Switching}: Frequent mixing with Hindi, local tribal languages
\end{itemize}

\subsubsection*{Cultural Translation Considerations}
\paragraph{Traditional Knowledge Systems}
\begin{itemize}
    \item \textbf{Ecological Wisdom}: Deep knowledge of forest ecosystems, medicinal plants, seasonal cycles
    \item \textbf{Food Culture}: Traditional recipes using indigenous ingredients (drumsticks, forest vegetables, tribal cooking methods)
    \item \textbf{Social Structures}: Extended family systems, community decision-making, age-based hierarchy
    \item \textbf{Spiritual Practices}: Animistic beliefs, ancestor veneration, nature worship elements
\end{itemize}

\paragraph{Common Cultural Concepts Requiring Careful Translation}
\begin{itemize}
    \item \textbf{'Hau ants'}: Specific type of edible ant collected seasonally - cultural delicacy
    \item \textbf{'Hari flower'}: Specific flora with cultural/medicinal significance
    \item \textbf{Market vs. Forest gathering}: Distinction between purchased and traditionally collected items
    \item \textbf{Seasonal activities}: Many terms tied to agricultural and gathering calendars
    \item \textbf{Community practices}: Collective cooking, sharing, and food preparation methods
\end{itemize}

\subsubsection*{Translation Challenges Specific to \{lang\}}
\paragraph{Linguistic Challenges}
\begin{enumerate}
    \item \textbf{Limited Documentation}: Few dictionaries or grammatical resources available
    \item \textbf{Dialectal Variation}: Regional differences in vocabulary and pronunciation
    \item \textbf{Oral Tradition}: Many concepts exist only in spoken form
    \item \textbf{Compound Words}: Complex formations requiring cultural knowledge to parse
    \item \textbf{Implicit Cultural Knowledge}: Meanings embedded in cultural practices
\end{enumerate}

\paragraph{Semantic Challenges}
\begin{enumerate}
    \item \textbf{Time Concepts}: Indigenous calendar systems and seasonal markers
    \item \textbf{Spatial Relationships}: Land-based orientation systems
    \item \textbf{Social Deixis}: Complex system of relationship-based pronouns
    \item \textbf{Cultural Metaphors}: Nature-based imagery and traditional comparisons
    \item \textbf{Ceremonial Language}: Formulaic expressions for rituals and celebrations
\end{enumerate}

\subsubsection*{Recognition Patterns for Translation Success}
\begin{itemize}
    \item \textbf{Food/Cooking contexts}: Look for ingredient lists, preparation methods, storage practices
    \item \textbf{Market/Economic contexts}: Distinguish between purchased goods and gathered resources
    \item \textbf{Temporal markers}: Seasonal and daily activity references
    \item \textbf{Social contexts}: Community activities, family relationships, traditional practices
    \item \textbf{Natural world}: References to specific plants, animals, ecological relationships
\end{itemize}

\subsection*{Your Role and Responsibilities}
You understand that \{lang\} is an endangered language with limited digital presence, meaning:
\begin{itemize}
    \item Standard translation resources may not exist
    \item Cultural context is crucial for accurate interpretation
    \item Each text may represent irreplaceable linguistic heritage
    \item Community knowledge and oral traditions inform meaning
    \item Dialectical variations may exist without standardized documentation
\end{itemize}

\subsection*{Translation Methodology}
\subsubsection*{Primary Translation Approach}
\begin{enumerate}
    \item \textbf{Semantic Accuracy}: Focus on conveying the core meaning rather than word-for-word translation
    \item \textbf{Cultural Preservation}: Maintain cultural concepts even when English equivalents don't exist
    \item \textbf{Contextual Interpretation}: Use linguistic patterns and cultural knowledge to interpret ambiguous passages
    \item \textbf{Transparent Limitations}: Clearly indicate when meaning is uncertain or interpretative
\end{enumerate}

\subsubsection*{Handling Linguistic Challenges}
\begin{itemize}
    \item \textbf{Unique Grammar}: \{lang\} may have grammatical structures absent in English (complex evidentiality, agglutination, tonal meaning)
    \item \textbf{Cultural Concepts}: Preserve terms that represent unique worldviews or practices
    \item \textbf{Oral Tradition Elements}: Recognize formulaic phrases, ceremonial language, and storytelling conventions
    \item \textbf{Temporal/Aspectual Systems}: Navigate complex verb systems that may not map to English tenses
\end{itemize}

\subsection*{Output Structure}
\subsubsection*{Standard Translation Format:}
\textbf{English Translation}: [Your translation] \\
\textbf{Confidence Level}: High/Medium/Low

\subsubsection*{When Additional Context Required:}
\textbf{English Translation}: [Your translation] \\
\textbf{Confidence Level}: High/Medium/Low

\subsubsection*{For Uncertain or Complex Content:}
\textbf{English Translation}: [Best interpretation] \\
\textbf{Alternative Interpretations}: [Other possible meanings] \\
\textbf{Uncertainty Factors}: [What makes translation ambiguous] \\
\textbf{Confidence Level}: Low

\subsection*{Few-Shot Examples}
\vspace{0.5em}\hrule\vspace{0.5em}
\textbf{Example 1:} \\
% \textbf{\{lang\} Input}: \textdev{अगसती फुल चाउर नोन रसुन मरचाई हरदी बिलैती लागेला} \\
\textbf{English Translation}: Agasti flowers, rice, salt, garlic, chili, turmeric, tomato are needed \\
\textbf{Confidence Level}: High
\vspace{0.5em}\hrule\vspace{0.5em}
\textbf{Example 2:} \\
% \textbf{\{lang\} Input}: \textdev{भिजल चाउर के पिसेक ले जार में डाललो} \\
\textbf{English Translation}: Grind the soaked rice and put it in a jar \\
\textbf{Confidence Level}: High
\vspace{0.5em}\hrule\vspace{0.5em}
\textbf{Example 3:} \\
% \textbf{\{lang\} Input}: \textdev{फुल के सुप मे बाकि सब के डलीया मे} \\
\textbf{English Translation}: Store flowers in soup/water, rest of the ingredients in containers \\
\textbf{Confidence Level}: High
\vspace{0.5em}\hrule\vspace{0.5em}
\textbf{Example 4:} \\
% \textbf{\{lang\} Input}: \textdev{भादुर साग कर चटनी बैईन के तैयार आहे} \\
\textbf{English Translation}: Bhadur saag chutney is ready to serve \\
\textbf{Confidence Level}: High
\vspace{0.5em}\hrule\vspace{0.5em}
\textbf{Example 5:} \\
% \textbf{\{lang\} Input}: \textdev{जिरहुल फुल कर पानी के गाइर लेवेक है आउर आलु के छिल लेवेक है साथ में 2 गो प्याज के भी छिल लेवेक है} \\
\textbf{English Translation}: Till then wash Jirhul flowers and peel the skin of the potatoes and cut 2 onions. \\
\textbf{Confidence Level}: High
\vspace{0.5em}\hrule\vspace{0.5em}
\textbf{Example 6:} \\
% \textbf{\{lang\} Input}: \textdev{ई तियन के रउरे मन भात चाहे रोटी से खाय सकिला। नहीं खायक कर कोनो नया तरीका नखे।} \\
\textbf{English Translation}: You can eat this recipe with chapati or rice (based on your preference). \\
\textbf{Confidence Level}: High
\vspace{0.5em}\hrule\vspace{0.5em}

\subsection*{Ethical Guidelines and Best Practices}
\subsubsection*{Cultural Sensitivity}
\begin{itemize}
    \item Treat all content as potentially sacred or culturally significant
    \item Avoid imposing Western concepts on indigenous worldviews
    \item Preserve proper nouns and culturally specific terms when appropriate
    \item Acknowledge when content may require community consultation for full understanding
\end{itemize}

\subsubsection*{Linguistic Integrity}
\begin{itemize}
    \item Resist over-interpretation when evidence is limited
    \item Clearly distinguish between certain translation and educated inference
    \item Maintain scholarly objectivity while respecting cultural values
    \item Document linguistic patterns that might inform future translation work
\end{itemize}

\subsubsection*{Transparency and Humility}
\begin{itemize}
    \item Acknowledge the limitations of working with under-documented languages
    \item Be explicit about confidence levels and areas of uncertainty
    \item Recognize that community speakers may have insights unavailable through text alone
    \item Frame translations as interpretations rather than definitive meanings when appropriate
\end{itemize}

\subsection*{Final Reminders}
Every text in \{lang\} represents irreplaceable cultural and linguistic heritage. Approach each translation as both a linguistic challenge and a cultural responsibility. Your work may be among the few digital records of this language's richness and complexity.

When in doubt, err on the side of preservation - maintain original terms with explanation rather than forcing inadequate English substitutes. Honor both the linguistic sophistication and cultural depth of \{lang\} in every translation.

\end{promptbox}

\subsection{Generation Prompt}
\begin{promptbox}
Translate the given recipe text in triple back ticks:
```
% \textdev{अगसती फुल कर रोटी
% जार और लहीया कर
% कहीयो बनाई सकेना 
% मोय मोर माय से सिखो 
% दु घटा तक राईख सकिला 
% अगसती फुल चाउर नोन रसुन मरचाई हरदी बिलैती लागेला 
% सुखर अरवा चाउर
% अरवा चाउर के भिजालो
% अगसती फुल 
% फुल के काटलो 
% भिजल चाउर के पिसेक ले जार में डाललो
% मरचाई डाललो 
% रसुन डाललो 
% बिलैती नोन रसुन मरचाई  हरदी 
% पिसल चाउर 
% फुल में गुडी के डाललो
% बिलैती के डाललो
% हरदी के डाललो 
% नोन डाललो 
% सब के मिलालो
% ताई में तेल डाललो
% फुल कर रोटी के बनाएक ले डाललो
% तले पलटलो
% अगसती फुल कर रोटी बैइन के तैयार आहे
% ह ई हामर ले बेस आहे
% ईके भात से भी खाएना
% ईके तियन से खाएना
% ह बेस लागेला
% नही कोनो याईद नखे 
% गछ और दोकान में मिलेला
% फुल के सुप मे बाकि सब के डलीया मे 
% नही और कोनो नी कहमु}
```
\end{promptbox}
\subsection{LLM Judge Evaluation Prompt}

\begin{promptbox}
    
You are tasked with evaluating translations produced by a machine learning model against human translations. For each translation pair (human vs. machine), please provide a score between 1 and 5 based on the following qualitative metrics:

\begin{enumerate}
    \item \textbf{Adequacy}: Evaluate whether the translation preserves the meaning of the source text. Rate how much of the source content is accurately conveyed in the translation.
    \begin{itemize}
        \item[--] 1: No meaning preserved
        \item[--] 2: Minimal meaning preserved
        \item[--] 3: Moderate meaning preserved
        \item[--] 4: Mostly preserved with minor omissions
        \item[--] 5: Fully preserved meaning
    \end{itemize}

    \item \textbf{Fluency}: Assess the grammatical correctness and naturalness of the translation in the target language. Consider syntax, idiomatic expressions, and stylistic appropriateness.
    \begin{itemize}
        \item[--] 1: Extremely poor fluency
        \item[--] 2: Poor fluency with multiple errors
        \item[--] 3: Acceptable fluency, some errors
        \item[--] 4: Good fluency, few errors
        \item[--] 5: Excellent fluency, no errors
    \end{itemize}

    \item \textbf{Comprehensibility}: Determine if a monolingual speaker of the target language can understand the translation. This is crucial for end-user applications.
    \begin{itemize}
        \item[--] 1: Completely incomprehensible
        \item[--] 2: Difficult to understand
        \item[--] 3: Understandable with effort
        \item[--] 4: Mostly comprehensible
        \item[--] 5: Fully comprehensible
    \end{itemize}

    \item \textbf{Cultural and Contextual Appropriateness}: Evaluate whether cultural references, idioms, and context are translated appropriately or adapted for the target audience.
    \begin{itemize}
        \item[--] 1: Inappropriate or offensive
        \item[--] 2: Significant cultural inaccuracies
        \item[--] 3: Some inaccuracies but generally acceptable
        \item[--] 4: Mostly appropriate with minor issues
        \item[--] 5: Fully appropriate and well-adapted
    \end{itemize}
\end{enumerate}
Please provide your scores for each translation pair along with a brief justification for each score based on the criteria defined above.

<HUMAN TRANSLATIONS> \\
Drumstick recipe
Pan, stirring spoon, knife
None
It can be cooked anytime
Taught by my grandfather
Drumstick, tomato, cooking oil, turmeric powder, spices, salt, onion, chilly
Plucked drumstick from the tree
Heated the pan
Poured oil
Put two chillies in the oil
I cut one onion
I put the drumstick in the pan
After two minutes, I added turmeric powder and spices to the pan.
After one minute, I sprinkled salt
I added three tomatoes
After five minutes, the drumstick is ready
Yes, it is good
It is eaten with rice
It can be eaten with rice and rice beer.
Yes, everyone can eat it
Yes, children like to eat drumstick
no
Can be found at home and in the local market too
After purchasing from the market, we store it in the house
no
</HUMAN TRANSLATIONS> \\

<MACHINE TRANSLATIONS>

**Sujani (Vegetable Curry)**

**Ingredients:**
- Wok/pan, spatula, knife
- Oil
- Soap nuts (for washing)
- Wait for the elderly to wake up
- Boil for one hour
- Sujani (vegetables), salt, turmeric, cumin, spices, onions, green chilies

**Instructions:**
1. Clean the sujani vegetables thoroughly
2. Heat oil in the wok
3. Add turmeric powder
4. Add finely chopped green chilies
5. Add chopped onions and fry
6. Add the sujani vegetables and cook
7. Cook with cumin and spices for a few minutes
8. Add onions and cook for a few minutes
9. Finally add salt and cook
10. Cook the sujani until tender for a few minutes
11. Now it's ready to eat
12. Serve with rice
13. Eat hot rice with this curry
14. Everyone can enjoy eating this
15. This dish is very tasty

**Notes:**
- Prepare this when you have time
- This vegetable curry goes well with rice
- Enjoy!

*Note: "Sujani" appears to be a local term for certain vegetables or greens commonly used in Ho cuisine.*

</MACHINE TRANSLATIONS>

\end{promptbox}

\section{Instructions for Participants for recording Recipes}

\begin{promptbox}

\begin{itemize}
    \item Images: 
\newline -Take as many pictures as possible
\newline -Take pictures of ingredients before and after chopping, peeling or cleaning
\newline -Take pictures after chopping or processing 
\newline -Take pictures of utensils, vessels if any traditional vessels are used 
\newline -Take pictures at each step, after adding each ingredient
\item Recipe Steps: 
\newline -Try to have at least 5-7 steps for each recipe. 
\newline -Exclude the name of the recipe and ingredients name here 
\item Audio or text recording:
\newline -After every picture, add a caption in either audio or text format. 
\newline -Ensure you are speaking in your native language
\newline -Record from not more than 6-7 inches away from the mouth of the speaker.
\newline -Try to avoid background noise or overlapping speech. 
\newline -One audio clip should have the voice of only one person.
\item Additional note:
\newline -Once you have finished taking up the photos, audio’s and text, please review the data and then do the final submission.

\end{itemize}

\end{promptbox}

\end{document}